\documentclass{article} 
\usepackage{iclr2026_conference}
\usepackage{times}

\usepackage{amsmath,amsfonts,bm}

\def\eqref#1{equation~\ref{#1}}

\def\1{\bm{1}}

\DeclareMathAlphabet{\mathsfit}{\encodingdefault}{\sfdefault}{m}{sl}
\SetMathAlphabet{\mathsfit}{bold}{\encodingdefault}{\sfdefault}{bx}{n}

\usepackage{hyperref}
\usepackage{url}
\usepackage{graphicx}
\usepackage{booktabs}
\usepackage{pgfplots}
\pgfplotsset{compat=1.18}

\title{Decodable But Not Detachable: Training 
Data Granularity Determines Parametric Modularity in Large Language Models}

\author{
  Marcus Armstrong \quad Navid Ayoobi \quad Arjun Mukherjee \\
  Department of Computer Science \\
  University of Houston \\
  Houston, TX 77204 \\
  \texttt{\{miarmstr, nyoobi\}@cougarnet.uh.edu, amukher6@central.uh.edu}
}

\iclrfinalcopy
\begin{document}

\maketitle

\begin{abstract}
Do large language models contain domain-specific parametric shells: 
concentrated, causally necessary neuron populations whose removal 
selectively degrades a target domain while sparing others? We apply 
a uniform causal methodology across two domain granularities, three 
model families (1.5B to 7B parameters), and eight domains. At the 
academic subject level, zero neurons exceed 60\% domain selectivity 
across 939,008 combined FFN neurons and causal damage matrices are 
flat, despite domain identity being linearly decodable above 85\% 
accuracy. At the language and modality level, 0.65--1.14\% of neurons 
exceed 60\% selectivity, damage matrices are near-perfectly diagonal 
(ratios up to 595:1), and shell neuron sets are essentially disjoint 
(IoU $< 0.003$). Masking code-selective neurons reduces mathematical 
reasoning accuracy by 16--24 percentage points across all models; 
masking Spanish or Chinese neurons leaves it at or below random. 
Shell strength increases monotonically with scale and shells are 
spatially interleaved in a pattern that precludes group-level selective 
quantization. Parametric shells form where and only where training 
data was modular at the token level.
\end{abstract}

\section{Introduction}

Large language models (LLMs) process dozens of languages and thousands of 
subject domains within a single set of parameters, yet how this capacity is 
parametrically organized remains poorly understood. Prior work has attempted 
to identify language-specific neurons in multilingual LLMs: \citet{tang2024lape} 
propose LAPE to locate neurons by differential activation likelihood, but 
\citet{le2026crane} subsequently show that activation preference does not 
imply functional necessity --- LAPE-identified neurons achieve LangSpec-F1 
near zero under direct causal intervention. \citet{le2026crane} address this 
with CRANE, a relevance-attribution framework with substantially stronger 
causal effects, but evaluate a single model family across three languages 
without a null condition against which to assess when parametric shells exist 
and when they do not.

This paper addresses that open question. We apply a uniform causal methodology 
across two domain granularities, three model families (1.5B to 7B parameters), 
and eight domains: four academic subject categories and four language and 
modality domains (English, code, Spanish, Chinese). For each domain, we capture 
mean pre-activation magnitudes across the FFN intermediate dimension, identify 
domain-selective neurons via a share-based metric with backbone exclusion, and 
measure causal damage matrices net of a count-matched random baseline. The 
results reveal a sharp granularity boundary: zero neurons exceed 60\% domain 
selectivity at the subject level across 939,008 combined FFN neurons, while 
0.65--1.14\% exceed this threshold at the language level with near-perfectly 
diagonal damage matrices and shell IoU below 0.003. Parametric shells form 
where and only where training data was modular at the token level.

We make the following contributions:

\begin{enumerate}
    \item \textbf{A granularity boundary for parametric modularity.} We 
    establish that language and modality domains produce functional parametric 
    shells while academic subject domains do not, despite equivalent linear 
    decodability, contextualizing prior work by identifying the conditions 
    under which shells form.

    \item \textbf{Shell concentration and functional load.} We provide the 
    first systematic Pareto analysis of language shell structure, showing 
    that 2\% of neurons carries approximately 94\% of maximum domain-specific 
    capability loss across all four language domains and all three model families.

    \item \textbf{Scale analysis of shell effects.} Diagonal dNLL increases 
    monotonically with model scale (1.82 to 3.37 nats across 1.5B to 7B 
    parameters), indicating that larger models develop more functionally 
    load-bearing shells independent of the proportion of selective neurons.

    \item \textbf{Functional and spatial characterization.} Code-selective 
    neurons carry mathematical reasoning capability across all tested models; 
    language-selective neurons concentrate in final network layers for 
    non-English domains; and functional shells are spatially interleaved 
    throughout the weight matrices, precluding group-level selective quantization.
\end{enumerate}

\section{Related Work}

\paragraph{Neuron specialization in language models.}
Prior work has established that individual FFN neurons encode interpretable 
features \citep{bau2018identifying}, that specific neurons correlate with 
factual associations and their suppression degrades factual recall 
\citep{dai2022knowledge}, and that the FFN intermediate dimension functions 
as a population of sparse feature detectors rather than a monolithic block 
\citep{voita2023neurons}. Subsequent work localizes task-specific 
\citep{zhang2023finding}, culture-specific \citep{song2024identifying}, 
and syntactic \citep{mueller2022coloring} neurons within this population. 
Our work applies the same causal methodology to ask not which domains have 
dedicated neurons but under what conditions any domain acquires them.

\paragraph{Language-specific neurons in multilingual LLMs.}
\citet{tang2024lape} identify language-specific neurons via activation 
entropy (LAPE) and show that deactivating them degrades target-language 
performance. However, \citet{le2026crane} demonstrate that activation 
preference does not imply functional necessity: LAPE-identified neurons 
achieve LangSpec-F1 near zero under direct causal intervention. 
\citet{le2026crane} address this with CRANE, a relevance-attribution 
framework with substantially stronger causal effects, though limited to 
a single model family and three languages without a null condition. 
\citet{kojima2024language} and \citet{gurgurov2025language} report 
complementary findings, noting that language-selective neuron sets are 
nearly disjoint and sufficient to steer output language. Our work differs 
in three respects: we apply a uniform methodology across two domain 
granularities, producing the controlled null result that contextualizes 
the language-positive finding; we provide Pareto and scale analyses 
quantifying shell concentration and growth; and our causal effect 
magnitudes are substantially larger, with diagonal dNLL of 1.82--3.37 
nats versus the sub-unit perplexity changes reported by \citet{tang2024lape}.

\paragraph{Multilingual representation structure.}
\citet{wendler2024llamas} show that LLMs process non-English inputs by 
mapping them toward English representational space in intermediate layers 
before projecting back to the target language at output, providing a 
mechanistic account of the U-shaped layer distribution of language-specific 
neurons reported by \citet{tang2024lape}. Our layer distribution findings 
complicate this picture: in instruction-tuned models, language-selective 
neurons concentrate primarily in final layers rather than at both ends, 
consistent with instruction tuning absorbing input-side language mapping 
into the shared representation.

\section{Methodology}

\subsection{Models and Domains}

We evaluate three instruction-tuned decoder-only language models spanning 
distinct architectures and scales: Qwen2.5-1.5B-Instruct \citep{qwen2025}, 
Llama-3.2-3B-Instruct \citep{meta2024llama3}, and Mistral-7B-Instruct-v0.3 
\citep{jiang2023mistral}. These models contain $N = 250{,}880$, $229{,}376$, 
and $458{,}752$ FFN intermediate neurons respectively, for a combined total 
of $939{,}008$ neurons across all three model families.

We study two domain granularities. \textbf{Subject domains} consist of four 
labeled categories derived from MMLU \citep{hendrycks2021mmlu}, ARC-Challenge 
\citep{clark2018arc}, and OpenBookQA \citep{mihaylov2018obqa}: quantitative 
reasoning (abstract algebra, mathematics, physics, formal logic, and related 
subjects), biomedical (anatomy, biology, chemistry, clinical knowledge, and 
related subjects), humanities and social sciences (history, law, philosophy, 
economics, and related subjects), and elementary science (all ARC-Challenge 
and OpenBookQA items). Evaluation uses letter-logprob multiple-choice scoring. 
\textbf{Language and modality domains} consist of four corpora: English 
(WikiText-2 \citep{merity2017pointer}), code (MBPP \citep{austin2021mbpp}), 
Spanish, and Chinese (both from OPUS-100 \citep{tiedemann2020opus}). 
Evaluation uses mean negative log-likelihood (NLL) over 512-token segments. 
We additionally evaluate mathematical reasoning using GSM8K \citep{cobbe2021gsm8k} 
to test whether domain-selective neurons carry generalizable capability.

\subsection{Neuron Activation Capture}

For each domain $d$ and each model layer $l \in \{1, \ldots, L\}$, we 
capture the mean absolute pre-activation of each FFN intermediate neuron. 
Concretely, for a forward pass on input $x$, let 
$\mathbf{a}^{(l)}(x) \in \mathbb{R}^{d_f}$ denote the input to the 
down-projection matrix $\mathbf{W}^{(l)}_{\text{down}} \in 
\mathbb{R}^{d_h \times d_f}$ at layer $l$, averaged over all token 
positions. We define the mean activation of neuron $n = (l, j)$ for 
domain $d$ as:

\begin{equation}
    M_{d,n} = \mathbb{E}_{x \sim \mathcal{D}_d} 
    \left[ \frac{1}{T} \sum_{t=1}^{T} 
    \left| a^{(l)}_j(x)_t \right| \right]
\end{equation}

where $\mathcal{D}_d$ is the set of inputs for domain $d$ and $T$ is the 
sequence length. For subject domains, inputs consist of the question text 
alone without answer choices, computed via a mean-pooled forward pass. 
Omitting answer choices is essential: a pilot experiment using full 
multiple-choice prompts revealed that final-token representations cluster 
by the model's predicted answer letter rather than question content, 
confounding domain identification. For language domains, inputs are 
512-token text segments. All captures use a maximum input length of 1024 
tokens with item-level norm logging; items in the bottom 2\% by 
$\ell_2$ norm of the resulting representation are excluded from subspace 
fitting to prevent degenerate near-zero vectors from distorting domain 
statistics \citep[c.f.][]{armstrong2026tda}.

\subsection{Domain Selectivity}

Let $K$ denote the number of domains and $N = L \cdot d_f$ the total 
number of FFN neurons. We define the \textit{domain share} of neuron 
$n$ for domain $d$ as:

\begin{equation}
    \sigma_{d,n} = \frac{M_{d,n}}{\sum_{d'=1}^{K} M_{d',n} + \epsilon}
\end{equation}

Under uniform domain activity, $\sigma_{d,n} = 1/K$ for all $d$. The 
\textit{max-share} of neuron $n$ is $s_n = \max_d \sigma_{d,n}$, which 
equals $1/K$ for a neuron with no domain preference and approaches 1 for 
a neuron active exclusively for one domain. We report the distribution of 
$s_n$ across all neurons as the selectivity histogram, using $s_n > 0.60$ 
as our primary threshold for identifying neurons with strong domain 
preference.

We exclude a backbone set $\mathcal{B}$ consisting of the top 0.1\% of 
neurons by global mean activation $\bar{M}_n = \frac{1}{K}\sum_d M_{d,n}$ 
from all subsequent analyses. These neurons exhibit high activity across 
all domains and correspond to the massive-activation or super-weight 
phenomenon documented in prior work \citep{sun2024massive}; including 
them inflates apparent domain damage in masking experiments because their 
removal degrades all domains uniformly rather than selectively.

For each domain $d$, we define the domain shell $\mathcal{S}_d$ as the 
top-$m$ neurons ranked by excess share $\sigma_{d,n} - 1/K$, excluding 
$\mathcal{B}$, where $m$ is determined by a fixed mask fraction 
$\rho \in [0, 1]$ applied to the total neuron count:

\begin{equation}
    \mathcal{S}_d = \underset{n \notin \mathcal{B}}{\text{top-}m} 
    \left( \sigma_{d,n} - \frac{1}{K} \right), \quad m = \lfloor \rho N \rfloor
\end{equation}

We use $\rho = 0.02$ as our primary mask fraction and report Pareto 
curves over $\rho \in \{0.005, 0.01, 0.02, 0.05, 0.10, 0.20\}$. A 
count-matched random baseline $\mathcal{S}_{\text{rand}}$ is sampled 
uniformly from $\{1, \ldots, N\} \setminus \mathcal{B}$ at each 
fraction.

\subsection{Causal Masking and Damage Measurement}

To assess functional necessity, we intervene on the model by zeroing 
the intermediate activations at positions in a target shell during 
inference. Specifically, for a shell $\mathcal{S}_{d'}$ and an 
input $x$, the masked forward pass sets:

\begin{equation}
    \tilde{a}^{(l)}_j(x) = 
    \begin{cases} 
        0 & \text{if } (l, j) \in \mathcal{S}_{d'} \\
        a^{(l)}_j(x) & \text{otherwise}
    \end{cases}
\end{equation}

for all layers $l$ and all token positions. This is implemented via 
registered forward hooks on the down-projection pre-activation without 
modifying model weights.

For language domains, we measure the NLL change under masking. The 
damage of masking domain $d'$'s shell on language $d$ is reported 
net of the random baseline to isolate shell-specific effects from 
general disruption:

\begin{equation}
    \Delta_{d' \to d} = 
    \underbrace{\text{NLL}_d(\mathcal{S}_{d'})}_{\text{shell mask}} - 
    \underbrace{\text{NLL}_d(\mathcal{S}_{\text{rand}})}_{\text{random mask}}
\end{equation}

This yields a $K \times K$ damage matrix where diagonal entries 
indicate how much masking domain $d$'s shell degrades domain $d$ 
beyond random, and off-diagonal entries indicate cross-domain 
interference. For subject domains, accuracy replaces NLL as the 
evaluation metric and the same net-of-random normalization applies.

We summarize the damage matrix by its diagonal mean 
$\bar{\Delta}_{\text{diag}} = \frac{1}{K} \sum_d \Delta_{d \to d}$ 
and off-diagonal mean 
$\bar{\Delta}_{\text{off}} = \frac{1}{K(K-1)} \sum_{d \neq d'} 
\Delta_{d' \to d}$, and report their ratio as a single scalar measure 
of shell selectivity.

\subsection{Shell Overlap and Spatial Organization}

To assess the disjointness of domain shells, we compute the Jaccard 
similarity (IoU) between the top-$m$ shell sets for each domain pair:

\begin{equation}
    \text{IoU}(d, d') = 
    \frac{|\mathcal{S}_d \cap \mathcal{S}_{d'}|}
    {|\mathcal{S}_d \cup \mathcal{S}_{d'}|}
\end{equation}

To assess spatial organization, we record the layer-wise distribution 
of shell neurons by counting, for each layer $l$ and domain $d$, the 
number of neurons $(l, j) \in \mathcal{S}_d$.

To assess whether functional shell structure permits selective 
quantization, we apply per-group asymmetric INT4 fake quantization 
\citep{frantar2023gptq} to the down-projection weight matrix 
$\mathbf{W}^{(l)}_{\text{down}}$ under three conditions: uniform 
quantization of all weight groups, quantization of groups whose 
constituent neurons are entirely in $\bigcup_d \mathcal{S}_d^c$ 
(the core), and the inverse. Group size is $g = 128$ following 
standard deployment practice \citep{lin2024awq}. NLL is evaluated 
after weight replacement without any calibration data, isolating 
the native quantization compatibility of the weight configuration.

\section{Results}

\subsection{Subject-Level Domains: Zero Parametric Modularity}
\label{sec:subject}

\begin{figure}[t]
    \centering
    \includegraphics[width=\textwidth]{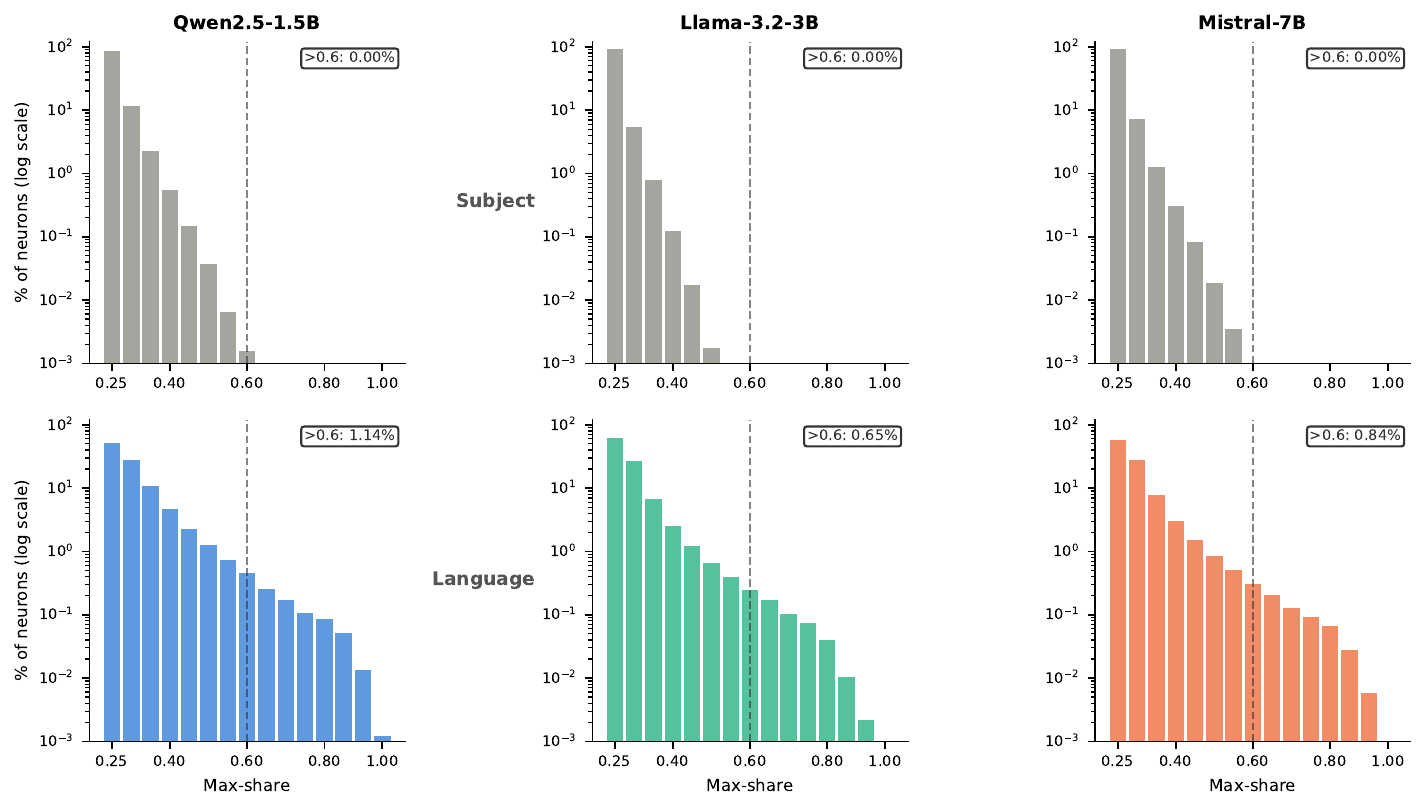}
    \caption{Selectivity histograms for subject-level domains (top row)
    and language/modality domains (bottom row) across three model families.
    The y-axis is log-scaled percentage of total FFN neurons. The dashed
    vertical line marks the 0.60 max-share threshold. At the subject level,
    no neurons exceed 0.60 across 939,008 combined neurons. At the language
    level, 0.65--1.14\% exceed 0.60 in all three models.}
    \label{fig:selectivity}
\end{figure}

Figure~\ref{fig:selectivity} (top row) shows the selectivity histogram
for subject-level domains across all three models. The distribution is
unimodal and rapidly decaying in all cases: no neuron exceeds 0.60
max-share across a combined 939,008 FFN neurons, and the 99th
percentile reaches at most 0.388.

Causal masking confirms functional indifference. The subject-level
damage matrix has a diagonal mean of 0.121--0.181 accuracy points and
an off-diagonal mean of 0.031--0.076, with no model exceeding a
diagonal-to-off-diagonal ratio of 2.4. The elem\_sci row is an
exception whose off-diagonal entries reveal general disruption rather
than a selective shell. These results hold despite domain identity
being linearly decodable above 85\% accuracy via a logistic regression
probe, providing a clean dissociation: domain information is present
in activation geometry but not supported by a dedicated parametric
substrate.

\subsection{Language-Level Domains: Concentrated Parametric Shells}
\label{sec:language}

Figure~\ref{fig:selectivity} (bottom row) shows the selectivity
histogram for language and modality domains. The fraction of neurons
exceeding 0.60 max-share is 1.14\%, 0.65\%, and 0.84\% for
Qwen2.5-1.5B, Llama-3.2-3B, and Mistral-7B respectively, compared
to 0.00\% in all three models at the subject level.

\begin{figure}[t]
    \centering
    \includegraphics[width=\textwidth]{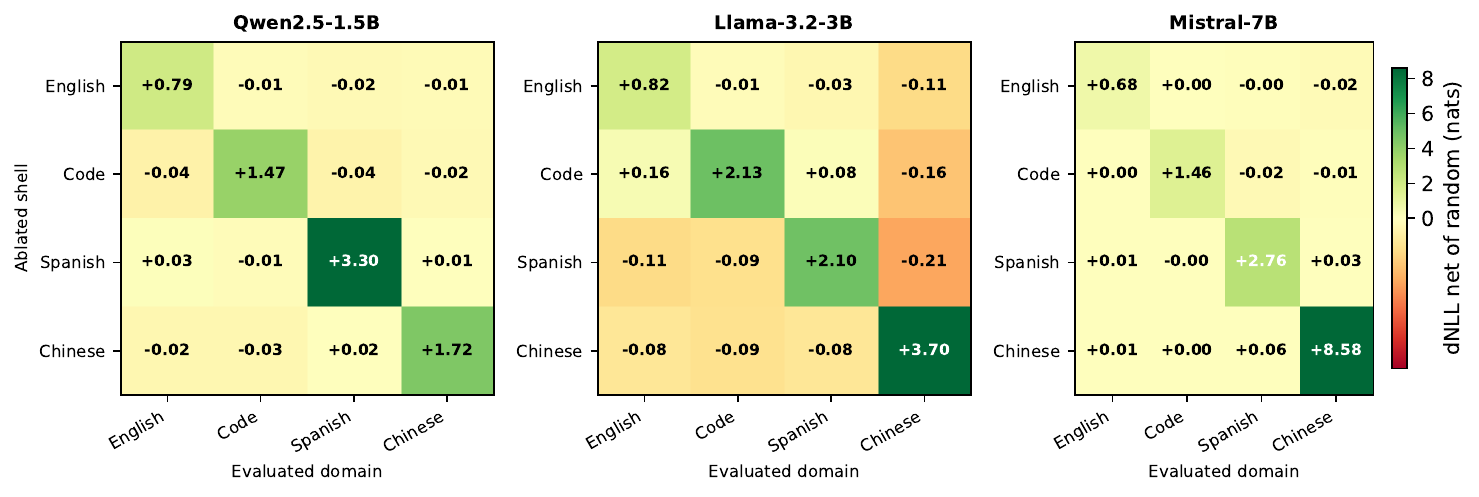}
    \caption{Language-level damage matrices (dNLL net of count-matched
    random baseline) for all three models at $\rho = 0.02$.
    Rows: ablated shell. Columns: evaluated domain. Diagonal means
    are 1.82, 2.19, and 3.37 nats; off-diagonal means are
    $-0.012$, $-0.060$, and $+0.006$ nats.}
    \label{fig:damage}
\end{figure}

Figure~\ref{fig:damage} shows the language-level damage matrices.
The matrices are strongly diagonal across all model families.
Diagonal means are $\bar{\Delta}_{\text{diag}} = 1.82$, $2.19$,
and $3.37$ nats; off-diagonal means are $-0.012$, $-0.060$, and
$+0.006$ nats. The diagonal-to-off-diagonal ratio ranges from 37:1
to above 500:1. The largest single entry is the Chinese diagonal in
Mistral-7B at $+8.58$ nats, against an English off-diagonal of
$+0.013$ nats under the same mask.

Shell disjointness is near-complete: the maximum IoU between any
two domain shell sets is 0.0029, with most pairs below 0.001
(Appendix~\ref{app:overlap}). Layer-wise shell distributions
(Appendix~\ref{app:layers}) show Spanish and Chinese neurons
concentrating in final network layers, while English and code
neurons distribute uniformly, consistent with instruction tuning
consolidating non-English language-specific computation into the
vocabulary projection stage.

\subsection{Pareto Analysis: Shell Concentration and Functional Load}
\label{sec:pareto}

\begin{figure}[t]
    \centering
    \includegraphics[width=\textwidth]{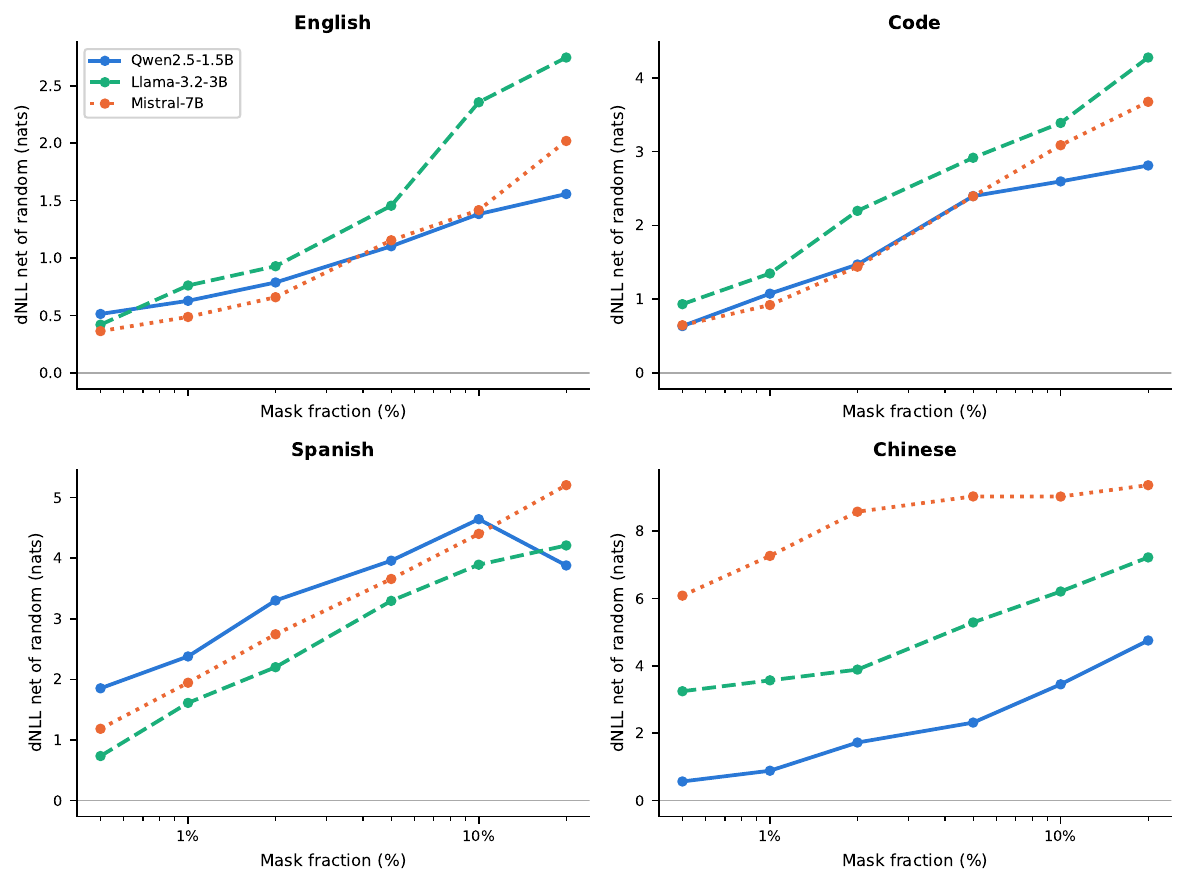}
    \caption{Diagonal dNLL net of random as a function of mask fraction 
    $\rho$ for each language domain across all three models (log-scaled 
    x-axis). Damage accumulates rapidly between $\rho = 0.005$ and 
    $\rho = 0.02$ and saturates by $\rho = 0.05$ for all domains. The 
    Chinese curve in Mistral-7B reaches $+6.09$ nats at $\rho = 0.005$ 
    (2,293 neurons) and $+8.58$ nats at $\rho = 0.02$, representing 
    91.6\% of the total damage observed at $\rho = 0.20$.}
    \label{fig:pareto}
\end{figure}

Figure~\ref{fig:pareto} shows damage accumulating rapidly between 
$\rho = 0.005$ and $\rho = 0.02$ and saturating by $\rho = 0.05$ 
for all language domains. At $\rho = 0.02$, damage represents on average 72\% of
the maximum observed; for Chinese in Mistral-7B, $\rho = 0.005$
(2,293 neurons) already produces $+6.09$ nats while the curve
saturates at $+9.03$ nats by $\rho = 0.05$. At the subject level,
damage curves are flat and undifferentiated from random throughout
the $\rho$ range.

\subsection{Code Shells Carry Mathematical Reasoning}
\label{sec:gsm8k}

Table~\ref{tab:gsm8k} reports GSM8K accuracy under each masking
condition net of the count-matched random baseline. Masking the
code shell produces the largest and most consistent effect:
$-17.3$, $-24.0$, and $-16.0$ percentage points for Qwen2.5-1.5B,
Llama-3.2-3B, and Mistral-7B respectively. Mathematical reasoning,
despite being expressed in English-language text, relies on the
code-selective neuron population rather than the English-selective
population.

Masking Spanish produces effects at or below random in all three
models. The Chinese result varies in a pattern consistent with
pretraining composition: Mistral-7B shows near-zero Chinese effect
($+0.7$pp), Qwen2.5-1.5B shows a small negative effect ($-1.3$pp),
and Llama-3.2-3B shows $-10.0$pp, tracing a gradient that directly
maps to presumed Chinese mathematical content during pretraining.
Shell purity is not a fixed property of a language but a property
of how that language was represented in the training data.\footnote{
Qwen2.5-1.5B's random mask condition nominally exceeds its unmasked
baseline (36.0\% vs. 30.7\%); this 1.4 standard error gap is
consistent with sampling variability. All Qwen effects are reported
net of this elevated baseline and are therefore conservative.}

\begin{table}[t]
\centering
\caption{GSM8K accuracy (\%) under each masking condition.
Net-of-random effects in parentheses. Bold: largest net
effect per model.}
\label{tab:gsm8k}
\begin{tabular}{lccc}
\toprule
Condition & Qwen2.5-1.5B & Llama-3.2-3B & Mistral-7B \\
\midrule
Baseline      & 30.7 & 37.3 & 40.7 \\
Random mask   & 36.0 & 29.3 & 34.0 \\
\midrule
mask\_english & 37.3 \scriptsize{(+1.3)}
              & 16.0 \scriptsize{($-$13.3)}
              & 32.0 \scriptsize{($-$2.0)} \\
mask\_code    & 18.7 \scriptsize{\textbf{($-$17.3)}}
              & 5.3  \scriptsize{\textbf{($-$24.0)}}
              & 18.0 \scriptsize{\textbf{($-$16.0)}} \\
mask\_spanish & 32.0 \scriptsize{($-$4.0)}
              & 29.3 \scriptsize{(0.0)}
              & 30.7 \scriptsize{($-$3.3)} \\
mask\_chinese & 34.7 \scriptsize{($-$1.3)}
              & 19.3 \scriptsize{($-$10.0)}
              & 34.7 \scriptsize{(+0.7)} \\
\bottomrule
\end{tabular}
\end{table}

\subsection{Scale Analysis}
\label{sec:scale}

\begin{figure}[h]
    \centering
    \includegraphics[width=0.85\textwidth]{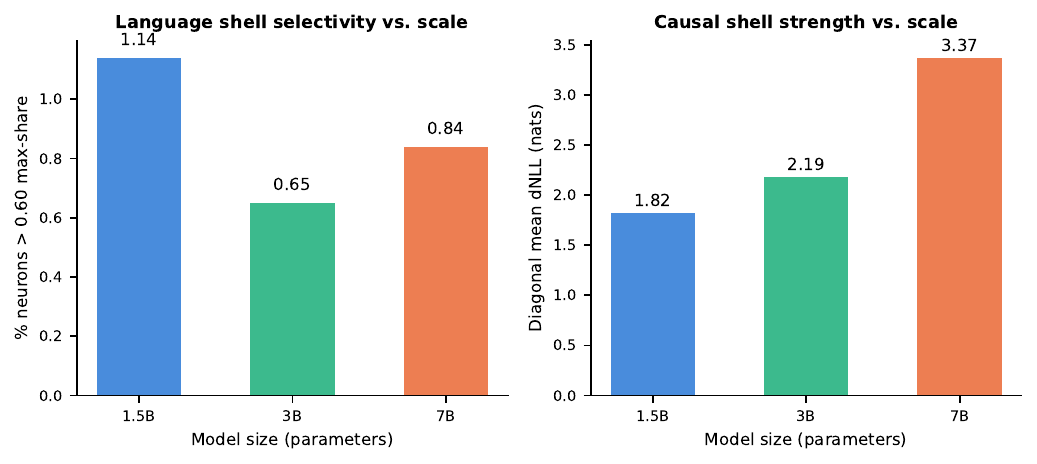}
    \caption{Language shell properties as a function of model scale. 
    Left: fraction of neurons exceeding 0.60 max-share shows a 
    non-monotonic relationship with parameter count. Right: diagonal 
    mean dNLL at $\rho = 0.02$ increases monotonically with scale 
    (1.82, 2.19, 3.37 nats), indicating that larger models develop 
    more functionally load-bearing language shells.}
    \label{fig:scale}
\end{figure}

Figure~\ref{fig:scale} shows language shell strength as a function 
of model scale. The fraction of neurons
exceeding 0.60 max-share is non-monotonic (1.14\%, 0.65\%, 0.84\%),
reflecting differences in architecture and pretraining composition.
In contrast, $\bar{\Delta}_{\text{diag}}$ increases monotonically:
1.82, 2.19, and 3.37 nats. At $\rho = 0.005$, Mistral-7B's Chinese
shell produces 6.09 nats of targeted damage versus 0.57 nats in
Qwen2.5-1.5B. Larger models develop not more selective neurons
proportionally, but more functionally load-bearing ones.

\subsection{Spatial Interleaving: Functional Shells Are
Parametrically Dispersed}
\label{sec:spatial}

Under per-group asymmetric INT4 quantization with group size
$g = 128$, a group qualifies for core-only quantization only if
all 128 neurons fall below the selectivity threshold. At shell
densities of 12--21\%, the probability that any group of 128
consecutive neurons is entirely core evaluates to
$2.2 \times 10^{-8}$ for the highest-density model. In practice,
zero qualifying groups are found in any of the three models.

The empirical consequence: int4\_core and int4\_inv produce NLL
values identical to BF16 to four decimal places. Uniform INT4
produces language-uniform increases of $+0.028$--$+0.081$ nats,
with no language showing systematically larger degradation,
consistent with quantization affecting only shared-core
representations. Functional modularity does not imply spatial
locality: language-selective neurons are interleaved throughout
the intermediate dimension at every spatial scale accessible to
standard group quantization.

\section{The Granularity Principle}

The results converge on a single organizing principle: parametric
shells form where and only where training data was itself modular
at the token level.

\paragraph{Why language domains produce shells.}
English, Spanish, Chinese, and code occupy largely non-overlapping
regions of the token distribution during pretraining. Gradient
updates for Chinese inputs systematically push a subset of
parameters toward configurations useful for Chinese processing
while leaving parameters used for English largely undisturbed.
Repeated across billions of tokens, this pressure produces a
population of neurons concentrated on one partition of the training
distribution. The shell IoU values below 0.003 are the parametric
residue of this partitioned training signal.

\paragraph{Why subject domains do not.}
A question about cellular respiration and a question about
differential equations are both written in English with similar
syntactic structures and overlapping academic vocabulary. The
gradient signal for biology questions and physics questions
updates largely overlapping parameter sets, producing no
systematic pressure toward domain-selective neurons. The 88\%
probe accuracy establishes that domain information is present in
activation geometry, but representational geometry and parametric
organization are separable properties of a trained network and
need not co-occur.

\paragraph{Shell purity tracks training data composition.}
Mistral-7B's Chinese shell carries essentially no mathematical
reasoning capability ($+0.7$pp on GSM8K net of random), while
Llama-3.2-3B's shows $-10.0$pp, and Qwen2.5-1.5B's shows
$-1.3$pp. This gradient maps directly to the presumed gradient
in Chinese mathematical content during pretraining. A shell
learned on purely linguistic content carries only linguistic
function; a shell learned on thematically diverse content
carries a corresponding mixture. Shell purity is a property
of the training data, not of the domain.

\section{Discussion}

\paragraph{Implications for multilingual model design.}
Parametric separability across languages is not an emergent property 
of scale: it is a direct consequence of whether training data partitions 
by token distribution. Languages whose training data overlaps substantially 
with English at the token level may develop weaker or less pure shells 
than typologically distant languages with distinct scripts. Deliberate 
curation of partition sharpness across target languages, rather than 
simply maximizing token counts, may produce more predictable behavior 
under language-targeted interventions.

\paragraph{Implications for editing and compression.}
For language domains, the concentrated and disjoint shell structure 
makes neuron-level editing feasible: the Chinese shell in Mistral-7B 
is identifiable, functionally load-bearing, and nearly orthogonal to 
all other domain shells. For subject domains, no analogous structure 
exists. For compression, spatial interleaving means functional 
modularity cannot be directly converted into memory efficiency under 
standard group-level quantization. Two directions remain open: 
per-neuron precision assignment, and weight-matrix reordering that 
clusters shell neurons contiguously before applying group quantization. 
Both are weight-equivalent and constitute concrete targets for 
future work.

\paragraph{Relationship to prior work.}
Our results extend \citet{tang2024lape} and \citet{le2026crane} in 
three respects. First, causal effect magnitudes are substantially 
larger: LAPE produces sub-unit perplexity changes and CRANE produces 
moderate LangSpec-F1 values, while our approach identifies neurons 
whose removal is functionally catastrophic, establishing that a 
concentrated population carries the functional load rather than 
merely correlating with language use. Second, the subject-level 
null condition, absent from prior work, is what makes the 
language-level positive result interpretable as a boundary 
condition rather than a data point. Third, the final-layer 
concentration we observe in instruction-tuned models differs 
from the U-shaped distribution reported for base models 
\citep{tang2024lape}, consistent with instruction tuning 
consolidating language-specific computation into the vocabulary 
projection stage.

\paragraph{Limitations.}
Our analysis covers three instruction-tuned model families in 
the 1.5B--7B range; whether the granularity principle holds for 
base models, larger models, or substantially different architectures 
is not established. The layer distribution comparison with 
\citet{tang2024lape} is indirect, as we evaluate different 
families under different training regimes. Our activation-magnitude 
identification method may not surface all functionally relevant 
neurons; gradient or relevance-based approaches may identify 
complementary populations. The subject-level null covers four 
coarse categories; finer-grained partitions or corpus-based 
taxonomies might reveal structure at intermediate granularities.

\section{Conclusion}

We investigated whether large language models contain domain-specific 
parametric shells: concentrated, causally necessary neuron populations 
whose removal selectively degrades a target domain while sparing others. 
Across two domain granularities, three model families, and eight domains, 
the answer depends entirely on granularity. At the subject level, zero 
neurons exceed 60\% domain selectivity across 939,008 combined FFN 
neurons and causal damage matrices are flat, despite domain identity 
being linearly decodable above 85\% accuracy. At the language and 
modality level, 0.65--1.14\% of neurons exceed 60\% selectivity, 
damage matrices are near-perfectly diagonal with ratios up to 595:1, 
shell neuron sets are essentially disjoint (IoU $< 0.003$), and 
masking code-selective neurons reduces mathematical reasoning accuracy 
by 16--24 percentage points while masking Spanish or Chinese neurons 
leaves it at or below random.

These results are unified by the granularity principle: parametric 
shells form where and only where training data was modular at the 
token level. The model carves its parameters the same way its training 
data was carved by script, vocabulary, and 
syntactic register.

\section*{AI Use Statement}
This manuscript was prepared with the assistance of large language model 
tools for drafting and editing. All experimental design, data collection, 
analysis, and scientific claims are the work of the authors. All 
AI-assisted text was reviewed, edited, and verified by the authors.

\section*{Ethics Statement}
This work analyzes publicly available open-weight language models 
(Qwen2.5-1.5B-Instruct, Llama-3.2-3B-Instruct, Mistral-7B-Instruct-v0.3) 
using publicly available benchmark datasets (MMLU, ARC-Challenge, OpenBookQA, 
GSM8K, WikiText-2, MBPP, OPUS-100). No human subjects were involved. No 
personal data was collected or used. Code will be released upon acceptance.

\section*{Reproducibility Statement}
All hyperparameters are documented in Table~\ref{tab:hyperparams} 
(Appendix~\ref{app:experimental}). Activation capture and causal masking 
implementations are described in detail in Appendix~\ref{app:experimental}. 
All experiments use a fixed random seed (42). Full model identifiers, 
dataset splits, and sampling procedures are specified in 
Section~\ref{sec:method} and Appendix~\ref{app:experimental}. 
Code will be released as supplementary material.

\bibliographystyle{abbrvnat}
\bibliography{references} 

@inproceedings{tang2024lape,
  title = "Language-Specific Neurons: The Key to Multilingual Capabilities in Large Language Models",
    author = "Tang, Tianyi  and
      Luo, Wenyang  and
      Huang, Haoyang  and
      Zhang, Dongdong  and
      Wang, Xiaolei  and
      Zhao, Xin  and
      Wei, Furu  and
      Wen, Ji-Rong",
    editor = "Ku, Lun-Wei  and
      Martins, Andre  and
      Srikumar, Vivek",
    booktitle = "Proceedings of the 62nd Annual Meeting of the Association for Computational Linguistics (Volume 1: Long Papers)",
    month = aug,
    year = "2024",
    address = "Bangkok, Thailand",
    publisher = "Association for Computational Linguistics",
    url = "https://aclanthology.org/2024.acl-long.309/",
    doi = "10.18653/v1/2024.acl-long.309",
    pages = "5701--5715"
}

@misc{le2026crane,
  title={CRANE: Causal Relevance Analysis of Language-Specific Neurons in Multilingual Large Language Models}, 
      author={Yifan Le and Yunliang Li},
      year={2026},
      eprint={2601.04664},
      archivePrefix={arXiv},
      primaryClass={cs.CL},
      url={https://arxiv.org/abs/2601.04664}, 
}

@inproceedings{dai2022knowledge,
  title = "Knowledge Neurons in Pretrained Transformers",
    author = "Dai, Damai  and
      Dong, Li  and
      Hao, Yaru  and
      Sui, Zhifang  and
      Chang, Baobao  and
      Wei, Furu",
    editor = "Muresan, Smaranda  and
      Nakov, Preslav  and
      Villavicencio, Aline",
    booktitle = "Proceedings of the 60th Annual Meeting of the Association for Computational Linguistics (Volume 1: Long Papers)",
    month = may,
    year = "2022",
    address = "Dublin, Ireland",
    publisher = "Association for Computational Linguistics",
    url = "https://aclanthology.org/2022.acl-long.581/",
    doi = "10.18653/v1/2022.acl-long.581",
    pages = "8493--8502"
}

@inproceedings{voita2023neurons,
  author = {Voita, Elena and Ferrando, Javier and Nalmpantis, Christoforos},
year = {2024},
month = {01},
pages = {1288-1301},
title = {Neurons in Large Language Models: Dead, N-gram, Positional},
doi = {10.18653/v1/2024.findings-acl.75}
}

@inproceedings{zhang2023finding,
  title = "Finding Skill Neurons in Pre-trained Transformer-based Language Models",
    author = "Wang, Xiaozhi  and
      Wen, Kaiyue  and
      Zhang, Zhengyan  and
      Hou, Lei  and
      Liu, Zhiyuan  and
      Li, Juanzi",
    editor = "Goldberg, Yoav  and
      Kozareva, Zornitsa  and
      Zhang, Yue",
    booktitle = "Proceedings of the 2022 Conference on Empirical Methods in Natural Language Processing",
    month = dec,
    year = "2022",
    address = "Abu Dhabi, United Arab Emirates",
    publisher = "Association for Computational Linguistics",
    url = "https://aclanthology.org/2022.emnlp-main.765/",
    doi = "10.18653/v1/2022.emnlp-main.765",
    pages = "11132--11152"
}

@inproceedings{song2024identifying,
  title = "Does Large Language Model Contain Task-Specific Neurons?",
    author = "Song, Ran  and
      He, Shizhu  and
      Jiang, Shuting  and
      Xian, Yantuan  and
      Gao, Shengxiang  and
      Liu, Kang  and
      Yu, Zhengtao",
    editor = "Al-Onaizan, Yaser  and
      Bansal, Mohit  and
      Chen, Yun-Nung",
    booktitle = "Proceedings of the 2024 Conference on Empirical Methods in Natural Language Processing",
    month = nov,
    year = "2024",
    address = "Miami, Florida, USA",
    publisher = "Association for Computational Linguistics",
    url = "https://aclanthology.org/2024.emnlp-main.403/",
    doi = "10.18653/v1/2024.emnlp-main.403",
    pages = "7101--7113"
}

@article{mueller2022coloring,
  title   = {Coloring the Black Box: What Syntactic Probes Learn},
  author  = {Mueller, Aaron and others},
  journal = {arXiv preprint},
  year    = {2022}
}

@article{kojima2024language,
  title   = {Language-Specific Neurons in Multilingual Models},
  author  = {Kojima, Takeshi and others},
  journal = {arXiv preprint},
  year    = {2024}
}

@article{gurgurov2025language,
  title   = {Targeted Fine-Tuning of Language-Specific Neurons
             in Multilingual Language Models},
  author  = {Gurgurov, Dmitry and others},
  journal = {arXiv preprint},
  year    = {2025}
}

@inproceedings{wendler2024llamas,
  title = "Do Llamas Work in {E}nglish? On the Latent Language of Multilingual Transformers",
    author = "Wendler, Chris  and
      Veselovsky, Veniamin  and
      Monea, Giovanni  and
      West, Robert",
    editor = "Ku, Lun-Wei  and
      Martins, Andre  and
      Srikumar, Vivek",
    booktitle = "Proceedings of the 62nd Annual Meeting of the Association for Computational Linguistics (Volume 1: Long Papers)",
    month = aug,
    year = "2024",
    address = "Bangkok, Thailand",
    publisher = "Association for Computational Linguistics",
    url = "https://aclanthology.org/2024.acl-long.820/",
    doi = "10.18653/v1/2024.acl-long.820",
    pages = "15366--15394"
}

@misc{qwen2025,
  title={Qwen2.5 Technical Report}, 
      author={Qwen and : and An Yang and Baosong Yang and Beichen Zhang and Binyuan Hui and Bo Zheng and Bowen Yu and Chengyuan Li and Dayiheng Liu and Fei Huang and Haoran Wei and Huan Lin and Jian Yang and Jianhong Tu and Jianwei Zhang and Jianxin Yang and Jiaxi Yang and Jingren Zhou and Junyang Lin and Kai Dang and Keming Lu and Keqin Bao and Kexin Yang and Le Yu and Mei Li and Mingfeng Xue and Pei Zhang and Qin Zhu and Rui Men and Runji Lin and Tianhao Li and Tianyi Tang and Tingyu Xia and Xingzhang Ren and Xuancheng Ren and Yang Fan and Yang Su and Yichang Zhang and Yu Wan and Yuqiong Liu and Zeyu Cui and Zhenru Zhang and Zihan Qiu},
      year={2025},
      eprint={2412.15115},
      archivePrefix={arXiv},
      primaryClass={cs.CL},
      url={https://arxiv.org/abs/2412.15115},
}

@misc{meta2024llama3,
  title={The Llama 3 Herd of Models}, 
      author={Aaron Grattafiori and Abhimanyu Dubey and Abhinav Jauhri and others},
      year={2024},
      eprint={2407.21783},
      archivePrefix={arXiv},
      primaryClass={cs.AI},
      url={https://arxiv.org/abs/2407.21783}, 
}

@misc{jiang2023mistral,
  title={Mistral 7B}, 
      author={Albert Q. Jiang and Alexandre Sablayrolles and Arthur Mensch and Chris Bamford and Devendra Singh Chaplot and Diego de las Casas and Florian Bressand and Gianna Lengyel and Guillaume Lample and Lucile Saulnier and Lélio Renard Lavaud and Marie-Anne Lachaux and Pierre Stock and Teven Le Scao and Thibaut Lavril and Thomas Wang and Timothée Lacroix and William El Sayed},
      year={2023},
      eprint={2310.06825},
      archivePrefix={arXiv},
      primaryClass={cs.CL},
      url={https://arxiv.org/abs/2310.06825}, 
}

@misc{hendrycks2021mmlu,
  title={Measuring Massive Multitask Language Understanding}, 
      author={Dan Hendrycks and Collin Burns and Steven Basart and Andy Zou and Mantas Mazeika and Dawn Song and Jacob Steinhardt},
      year={2021},
      eprint={2009.03300},
      archivePrefix={arXiv},
      primaryClass={cs.CY},
      url={https://arxiv.org/abs/2009.03300}, 
}

@misc{clark2018arc,
  title={Think you have Solved Question Answering? Try ARC, the AI2 Reasoning Challenge}, 
      author={Peter Clark and Isaac Cowhey and Oren Etzioni and Tushar Khot and Ashish Sabharwal and Carissa Schoenick and Oyvind Tafjord},
      year={2018},
      eprint={1803.05457},
      archivePrefix={arXiv},
      primaryClass={cs.AI},
      url={https://arxiv.org/abs/1803.05457}, 
}

@inproceedings{mihaylov2018obqa,
  title = "Can a Suit of Armor Conduct Electricity? A New Dataset for Open Book Question Answering",
    author = "Mihaylov, Todor  and
      Clark, Peter  and
      Khot, Tushar  and
      Sabharwal, Ashish",
    editor = "Riloff, Ellen  and
      Chiang, David  and
      Hockenmaier, Julia  and
      Tsujii, Jun{'}ichi",
    booktitle = "Proceedings of the 2018 Conference on Empirical Methods in Natural Language Processing",
    month = oct # "-" # nov,
    year = "2018",
    address = "Brussels, Belgium",
    publisher = "Association for Computational Linguistics",
    url = "https://aclanthology.org/D18-1260/",
    doi = "10.18653/v1/D18-1260",
    pages = "2381--2391"
}

@misc{merity2017pointer,
  title={Pointer Sentinel Mixture Models}, 
      author={Stephen Merity and Caiming Xiong and James Bradbury and Richard Socher},
      year={2016},
      eprint={1609.07843},
      archivePrefix={arXiv},
      primaryClass={cs.CL},
      url={https://arxiv.org/abs/1609.07843}, 
}

@misc{austin2021mbpp,
  title={Program Synthesis with Large Language Models}, 
      author={Jacob Austin and Augustus Odena and Maxwell Nye and Maarten Bosma and Henryk Michalewski and David Dohan and Ellen Jiang and Carrie Cai and Michael Terry and Quoc Le and Charles Sutton},
      year={2021},
      eprint={2108.07732},
      archivePrefix={arXiv},
      primaryClass={cs.PL},
      url={https://arxiv.org/abs/2108.07732},
}

@inproceedings{tiedemann2020opus,
  title = "{OPUS}-{MT} {--} Building open translation services for the World",
    author = {Tiedemann, J{\"o}rg  and
      Thottingal, Santhosh},
    editor = "Martins, Andr{\'e}  and
      Moniz, Helena  and
      Fumega, Sara  and
      Martins, Bruno  and
      Batista, Fernando  and
      Coheur, Luisa  and
      Parra, Carla  and
      Trancoso, Isabel  and
      Turchi, Marco  and
      Bisazza, Arianna  and
      Moorkens, Joss  and
      Guerberof, Ana  and
      Nurminen, Mary  and
      Marg, Lena  and
      Forcada, Mikel L.",
    booktitle = "Proceedings of the 22nd Annual Conference of the European Association for Machine Translation",
    month = nov,
    year = "2020",
    address = "Lisboa, Portugal",
    publisher = "European Association for Machine Translation",
    url = "https://aclanthology.org/2020.eamt-1.61/",
    pages = "479--480"
}

@misc{cobbe2021gsm8k,
  title={Training Verifiers to Solve Math Word Problems}, 
      author={Karl Cobbe and Vineet Kosaraju and Mohammad Bavarian and Mark Chen and Heewoo Jun and Lukasz Kaiser and Matthias Plappert and Jerry Tworek and Jacob Hilton and Reiichiro Nakano and Christopher Hesse and John Schulman},
      year={2021},
      eprint={2110.14168},
      archivePrefix={arXiv},
      primaryClass={cs.LG},
      url={https://arxiv.org/abs/2110.14168}, 
}

@article{armstrong2026tda,
  title   = {TITLE TO BE FILLED IN},
  author  = {Armstrong, Marcus and others},
  journal = {arXiv preprint},
  year    = {2026}
}

@inproceedings{sun2024massive,
  title={Massive Activations in Large Language Models},
author={Mingjie Sun and Xinlei Chen and J Zico Kolter and Zhuang Liu},
booktitle={First Conference on Language Modeling},
year={2024},
url={https://openreview.net/forum?id=F7aAhfitX6}
}

@misc{frantar2023gptq,
  title={GPTQ: Accurate Post-Training Quantization for Generative Pre-trained Transformers}, 
      author={Elias Frantar and Saleh Ashkboos and Torsten Hoefler and Dan Alistarh},
      year={2023},
      eprint={2210.17323},
      archivePrefix={arXiv},
      primaryClass={cs.LG},
      url={https://arxiv.org/abs/2210.17323},
}

@article{lin2024awq,
  author = {Lin, Ji and Tang, Jiaming and Tang, Haotian and Yang, Shang and Xiao, Guangxuan and Han, Song},
title = {AWQ: Activation-aware Weight Quantization for On-Device LLM Compression and Acceleration},
year = {2025},
issue_date = {December 2024},
publisher = {Association for Computing Machinery},
address = {New York, NY, USA},
volume = {28},
number = {4},
issn = {2375-0529},
url = {https://doi.org/10.1145/3714983.3714987},
doi = {10.1145/3714983.3714987},
journal = {GetMobile: Mobile Comp. and Comm.},
month = jan,
pages = {12–17},
numpages = {6}
}

@misc{bau2018identifying,
  title={Identifying and Controlling Important Neurons in Neural Machine Translation}, 
      author={Anthony Bau and Yonatan Belinkov and Hassan Sajjad and Nadir Durrani and Fahim Dalvi and James Glass},
      year={2018},
      eprint={1811.01157},
      archivePrefix={arXiv},
      primaryClass={cs.CL},
      url={https://arxiv.org/abs/1811.01157}, 
}

@inproceedings{wolf2020transformers,
  title = "Transformers: State-of-the-Art Natural Language Processing",
    author = "Wolf, Thomas  and
      Debut, Lysandre  and
      Sanh, Victor  and
      Chaumond, Julien  and
      Delangue, Clement  and
      Moi, Anthony  and
      Cistac, Pierric  and
      Rault, Tim  and
      Louf, Remi  and
      Funtowicz, Morgan  and
      Davison, Joe  and
      Shleifer, Sam  and
      von Platen, Patrick  and
      Ma, Clara  and
      Jernite, Yacine  and
      Plu, Julien  and
      Xu, Canwen  and
      Le Scao, Teven  and
      Gugger, Sylvain  and
      Drame, Mariama  and
      Lhoest, Quentin  and
      Rush, Alexander",
    editor = "Liu, Qun  and
      Schlangen, David",
    booktitle = "Proceedings of the 2020 Conference on Empirical Methods in Natural Language Processing: System Demonstrations",
    month = oct,
    year = "2020",
    address = "Online",
    publisher = "Association for Computational Linguistics",
    url = "https://aclanthology.org/2020.emnlp-demos.6/",
    doi = "10.18653/v1/2020.emnlp-demos.6",
    pages = "38--45"
}


\appendix

\section{Layer-wise Shell Distributions}
\label{app:layers}

\begin{figure}[h]
    \centering
    \includegraphics[width=\textwidth]{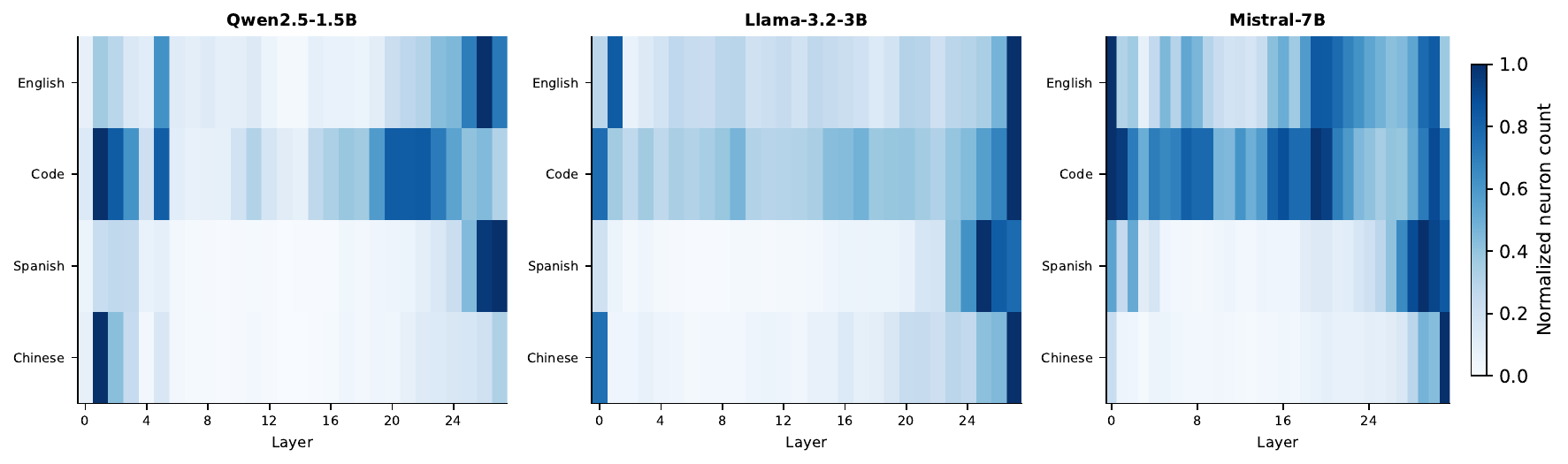}
    \caption{Layer-wise distribution of language shell neurons (normalized 
    by the maximum count across layers per domain) for each model. Spanish 
    and Chinese neurons concentrate strongly in final layers across all three 
    model families. English and code neurons are distributed more uniformly. 
    This final-layer concentration for non-English domains differs from the 
    U-shaped distribution reported for base models \citep{tang2024lape}, 
    consistent with instruction tuning absorbing input-side language mapping 
    into the shared representation.}
    \label{fig:layers}
\end{figure}

\section{Subject-Level Damage Matrices}
\label{app:subject_damage}

\begin{figure}[h]
    \centering
    \includegraphics[width=\textwidth]{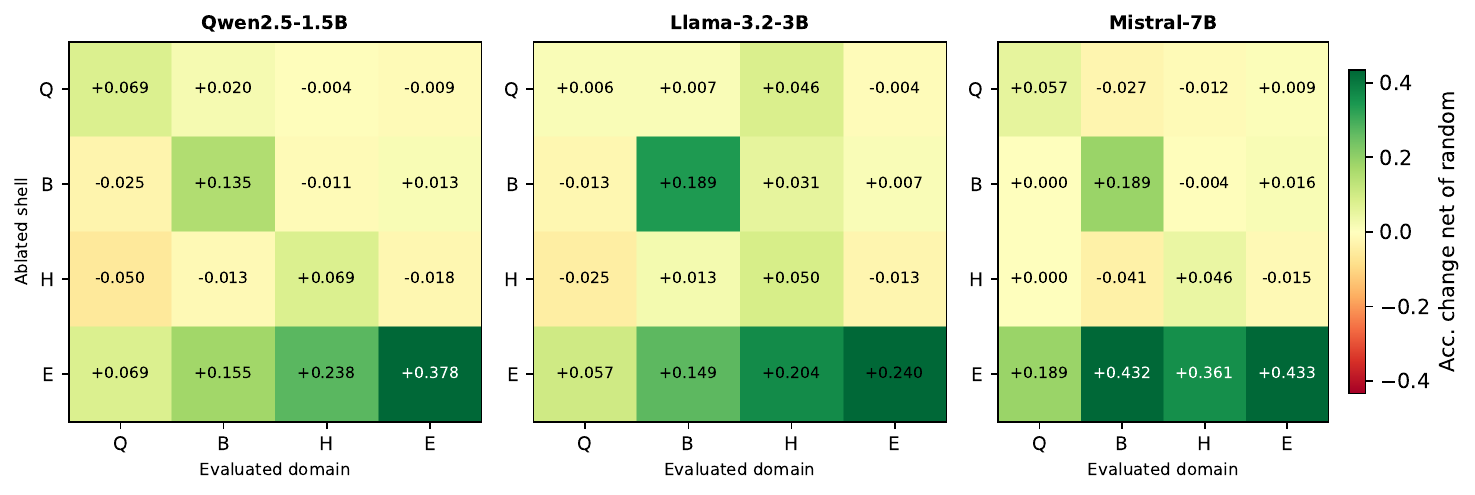}
    \caption{Subject-level damage matrices (accuracy points net of 
    random baseline) for all three models. Rows correspond to the 
    ablated shell; columns to the evaluated subject domain. Unlike 
    the language-level matrices, no model produces a consistently 
    diagonal structure. The elem\_sci row shows large off-diagonal 
    values consistent with disruption of shared general 
    question-answering capability rather than a domain-selective 
    shell. Domain labels: Q = quantitative, B = biomedical, H = 
    humanities and social sciences, E = elementary science.}
    \label{fig:app_subject_damage}
\end{figure}

Figure~\ref{fig:app_subject_damage} shows the subject-level damage 
matrices (net of count-matched random baseline, measured in accuracy 
points) for all three models. In contrast to the language-level 
matrices in Figure~\ref{fig:damage}, no model produces a 
consistently diagonal structure. For the quant, biomed, and humsoc 
rows, diagonal and off-diagonal entries are of comparable magnitude 
across all three models, consistent with the interpretation that 
masking any of these subject shells disrupts general question-answering 
capability rather than a domain-specific parametric substrate.

The elem\_sci row is an exception, showing larger values across all 
columns rather than a concentrated diagonal entry. This pattern 
reflects the composition of the elementary science evaluation set 
(ARC-Challenge and OpenBookQA), which draws on general English 
vocabulary and commonsense reasoning that overlaps substantially 
with all other subject domains. Masking the elem\_sci shell therefore 
degrades all domains, a signature of disrupting shared linguistic 
processing rather than a domain-specific shell. This interpretation 
is supported by the subject-level selectivity histogram (Figure 
\ref{fig:selectivity}, top row): the elem\_sci shell contains no 
neurons with max-share above 0.60 and therefore cannot constitute 
a parametrically selective substrate for elementary science in 
particular.

\section{Shell Overlap Matrices}
\label{app:overlap}

Tables~\ref{tab:overlap_language} and \ref{tab:overlap_subject} 
report the Jaccard similarity (IoU) between top-2\% shell sets 
for all domain pairs across all three models.

Language shell overlap is near-zero across all pairs and all models 
(Table~\ref{tab:overlap_language}). The maximum observed IoU is 
0.0029 (Mistral-7B, Spanish and Chinese), and the majority of 
pairs report IoU below 0.001. The slight elevation in the 
Spanish--Chinese pair relative to other pairs is consistent 
with the observation that Spanish and Chinese both require 
final-layer vocabulary mapping neurons that partially overlap 
in the intermediate dimension, though the overlap remains 
negligible in absolute terms.

Subject shell overlap is similarly near-zero across most pairs 
(Table~\ref{tab:overlap_subject}), with the exception of the 
biomed--elem\_sci pair, which reaches 0.0115--0.0162 across 
models. This modest elevation reflects the shared biological 
science content between biomedical MMLU questions and ARC 
elementary science questions. Even at its maximum, the overlap 
across all pairs and all models remains below 1.7\%, confirming 
that the absence of functional shells at the subject level is 
not attributable to excessive overlap between shell definitions.

\begin{table}[h]
\centering
\caption{Jaccard similarity (IoU) between top-2\% language shell 
sets for each domain pair and model. Values below 0.003 across 
all pairs and all models confirm near-complete shell disjointness.}
\label{tab:overlap_language}
\begin{tabular}{lcccc}
\toprule
Domain pair & Qwen2.5-1.5B & Llama-3.2-3B & Mistral-7B \\
\midrule
English -- Code    & 0.0001 & 0.0015 & 0.0001 \\
English -- Spanish & 0.0000 & 0.0004 & 0.0001 \\
English -- Chinese & 0.0001 & 0.0005 & 0.0001 \\
Code -- Spanish    & 0.0000 & 0.0001 & 0.0000 \\
Code -- Chinese    & 0.0000 & 0.0004 & 0.0001 \\
Spanish -- Chinese & 0.0000 & 0.0019 & 0.0029 \\
\bottomrule
\end{tabular}
\end{table}

\begin{table}[h]
\centering
\caption{Jaccard similarity (IoU) between top-2\% subject shell 
sets for each domain pair and model. The biomed--elem\_sci 
elevation reflects shared biological science content between 
MMLU biomedical questions and ARC elementary science items.}
\label{tab:overlap_subject}
\begin{tabular}{lccc}
\toprule
Domain pair & Qwen2.5-1.5B & Llama-3.2-3B & Mistral-7B \\
\midrule
Quant -- Biomed    & 0.0002 & 0.0009 & 0.0010 \\
Quant -- Humsoc    & 0.0000 & 0.0015 & 0.0005 \\
Quant -- Elem\_sci & 0.0005 & 0.0009 & 0.0014 \\
Biomed -- Humsoc   & 0.0003 & 0.0012 & 0.0013 \\
Biomed -- Elem\_sci & 0.0115 & 0.0136 & 0.0162 \\
Humsoc -- Elem\_sci & 0.0006 & 0.0012 & 0.0009 \\
\bottomrule
\end{tabular}
\end{table}

\section{Per-Model Pareto Curves}
\label{app:pareto_model}

\begin{figure}[h]
    \centering
    \includegraphics[width=\textwidth]{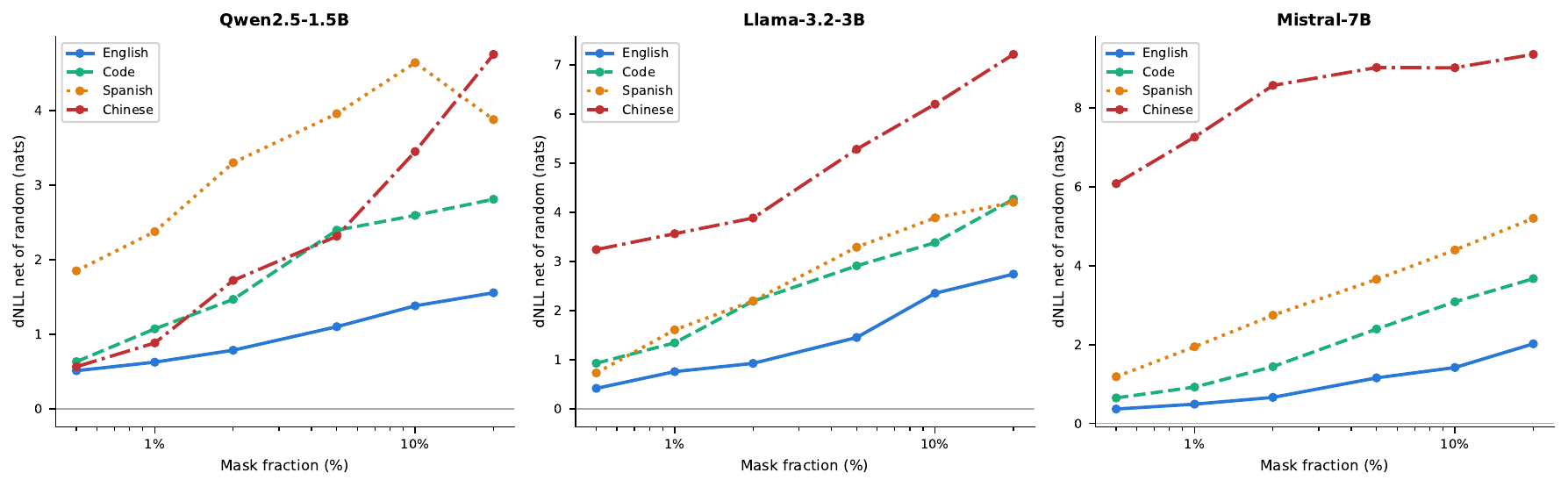}
    \caption{Diagonal dNLL net of random as a function of mask 
    fraction $\rho$, organized by model. Each panel shows all 
    four language domains for one model. Chinese and Spanish 
    curves are consistently steeper and saturate earlier than 
    English and code curves within each model, regardless of 
    architecture or scale.}
    \label{fig:app_pareto_model}
\end{figure}

Figure~\ref{fig:app_pareto_model} shows the Pareto curves 
reorganized by model rather than by language domain, allowing 
direct comparison of shell concentration across domains within 
each model. Within each model, Chinese and Spanish consistently 
show the steepest initial curves, saturating earliest. English 
shows the shallowest curve in all three models, consistent with 
English capability being distributed throughout the network 
rather than concentrated in a dedicated shell. The ordering 
of domain steepness is consistent across all three models 
despite differences in absolute dNLL magnitude, supporting 
the interpretation that shell concentration is a property 
of the domain partition rather than of any particular model 
architecture.

\section{Experimental Details}
\label{app:experimental}

\paragraph{Models.} All models are loaded in bfloat16 precision 
via HuggingFace Transformers \citep{wolf2020transformers} with 
\texttt{device\_map=auto}. Full model identifiers are 
\texttt{Qwen/Qwen2.5-1.5B-Instruct}, 
\texttt{meta-llama/Llama-3.2-3B-Instruct}, and 
\texttt{mistralai/Mistral-7B-Instruct-v0.3}.

\paragraph{Hardware.} All experiments were conducted on NVIDIA 
A100 80GB GPUs via Google Colab Pro. Qwen2.5-1.5B and 
Llama-3.2-3B experiments are compatible with a single T4 
16GB GPU. Mistral-7B requires at minimum an A100 40GB GPU 
in bfloat16 precision.

\paragraph{Dataset construction.} Subject domain evaluation 
pools are constructed from MMLU validation splits (600 items 
total, stratified across all available subjects with a minimum 
of 1 item per subject), ARC-Challenge test split (225 items, 
randomly sampled), and OpenBookQA test split (225 items, 
randomly sampled). Subject assignment follows the taxonomy 
in Section~\ref{sec:method}, with all sampling using a 
fixed random seed. Language domain corpora use 96 segments 
of 512 tokens each, drawn from WikiText-2 test split 
(English), MBPP test split (code), and OPUS-100 training 
splits (Spanish, Chinese) with streaming access. 
GSM8K evaluation uses 150 randomly sampled test items 
with greedy decoding and a maximum of 256 new tokens.

Table~\ref{tab:hyperparams} summarizes all primary 
hyperparameters used across experiments.

\begin{table}[h]
\centering
\caption{Experimental hyperparameters.}
\label{tab:hyperparams}
\begin{tabular}{ll}
\toprule
Parameter & Value \\
\midrule
Model precision          & bfloat16 \\
Activation capture depth & 0.75 (relative) \\
Norm exclusion threshold & bottom 2\% by $\ell_2$ norm \\
Backbone exclusion       & top 0.1\% by global mean activation \\
Primary mask fraction $\rho$ & 0.02 \\
Pareto fractions         & \{0.005, 0.01, 0.02, 0.05, 0.10, 0.20\} \\
INT4 group size          & 128 \\
INT4 calibration data    & none (calibration-free) \\
Max input length         & 1024 tokens \\
GSM8K decoding           & greedy, max 256 new tokens \\
Random seed              & 42 \\
Subject pool size        & 1017 items (600 MMLU + 225 ARC + 225 OBQA) \\
Language segments        & 96 $\times$ 512 tokens per domain \\
GSM8K pool size          & 150 items \\
\bottomrule
\end{tabular}
\end{table}

\paragraph{Activation capture implementation.}
Neuron activations are captured via PyTorch forward hooks 
registered on the \texttt{down\_proj} pre-activation at 
each FFN layer. For subject domains, hooks are applied 
to question-only forwards without answer choices or the 
\texttt{Answer:} scaffold token, and activations are 
mean-pooled across all non-padding token positions. For 
language domains, hooks are applied to teacher-forced 
forwards over fixed 512-token segments. Hook registration 
and removal are performed immediately before and after 
each forward pass to prevent cross-example contamination. 
All captures accumulate into float64 accumulators to 
prevent precision loss during averaging over large item 
counts.

\paragraph{Causal masking implementation.}
Neuron masking is implemented via the same forward hook 
mechanism, zeroing the pre-activation of selected neurons 
at all token positions across all forward passes in the 
evaluation set. Masks are defined once per condition and 
reused across all items; no per-item mask computation 
occurs at evaluation time. The random baseline mask is 
sampled once per experiment with a fixed random seed 
and held constant across all model and domain comparisons 
to ensure that random baseline variation does not 
confound cross-condition comparisons.

\end{document}